\documentclass{article}

\PassOptionsToPackage{numbers, compress}{natbib}
\usepackage[final]{nips_2016}
\usepackage{natbib}
\usepackage[utf8]{inputenc} 
\usepackage[T1]{fontenc}    
\usepackage{hyperref}       
\usepackage{url}            
\usepackage{booktabs}       
\usepackage{amsfonts}       
\usepackage{nicefrac}       
\usepackage{microtype}      
\usepackage{graphicx}
\usepackage{amsmath}
\usepackage{bm}
\usepackage{graphicx}
\usepackage{caption}
\usepackage{subcaption}

\title{Skin Cancer Detection and Tracking\\using Data Synthesis and Deep Learning}

\author{
  Yunzhu Li\thanks{indicate equal contribution.} \\
  Peking University\\
  \texttt{leo.liyunzhu@pku.edu.cn} \\
  \And
  Andre Esteva$^*$ \\
  Stanford University \\
  \texttt{esteva@cs.stanford.edu} \\
  \And
  Brett Kuprel \\
  Stanford University \\
  \texttt{brkuprel@gmail.com} \\
  \AND
  Rob Novoa \\
  Stanford University \\
  \texttt{ranovoa@gmail.com} \\
  \And
  Justin Ko \\
  Stanford University \\
  \texttt{jmko@stanford.edu} \\
  \And
  Sebastian Thrun \\
  Stanford University \\
  \texttt{thrun@stanford.edu}
}

\begin{document}

\maketitle



\begin{figure}[h!]
  \centering
  \includegraphics[width=0.7\textwidth]{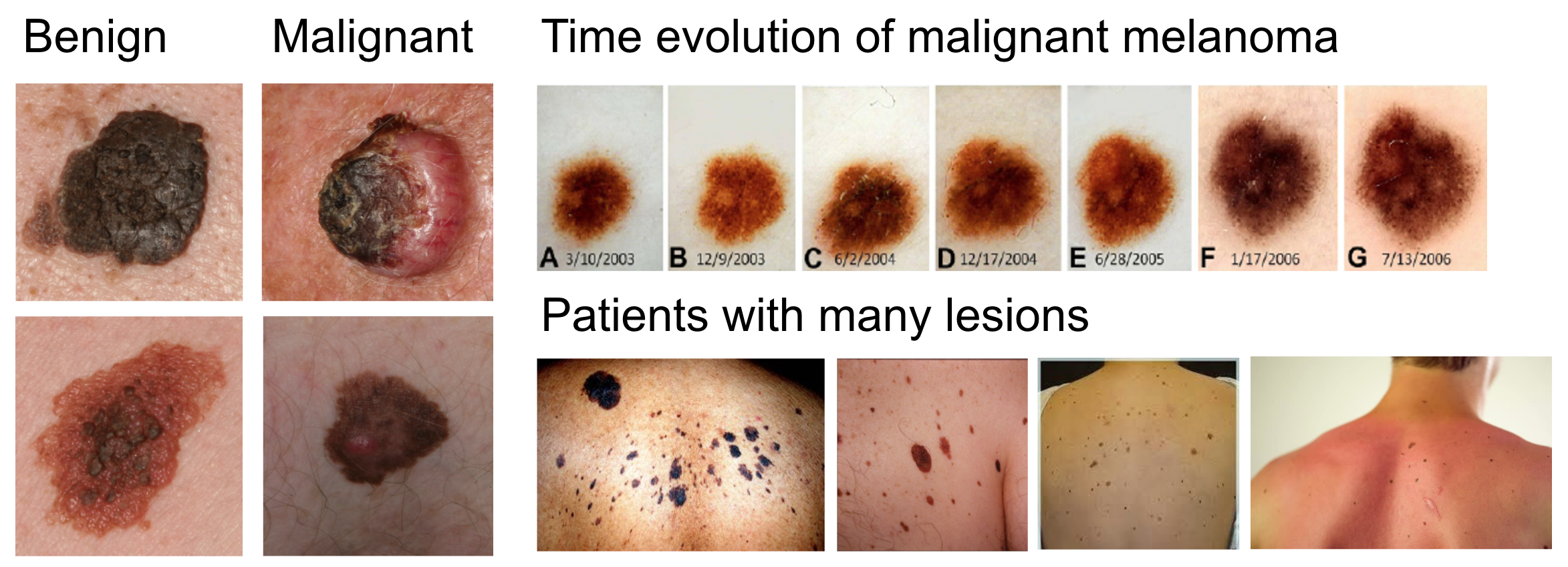}
  \caption{Key factors for skin cancer care include early detection and tracking over time.}
  \label{fig:pull}
\end{figure}

\section{Introduction}

Dermatology is a medical field which stands to be heavily augmented by the use of artificial intelligence techniques. 
Diseases are visually screened for, and many disease diagnoses are performed strictly with an in-clinic visual examination. 
Discerning between skin lesions is difficult - the difference between skin cancer (melanoma, carcinoma) and benign lesions (nevi, seborrheic keratosis) is minute (Figure~\ref{fig:pull}). 
With 5.4 million cases of skin cancer diagnosed each year in the United States alone, the need for quick and effective clinical screenings is rising ~\cite{rogers2015incidence,stern2010prevalence}.
Patients with skin cancer tend to be afflicted with many moles, and so one of the challenges in skin cancer screenings is identifying them amongst a myriad of benign lesions.
Another key element of these diagnoses is based on inspecting temporal changes in lesions - a fast changing lesion is more likely to be malignant. 
As such, patients and providers need tools to support this at scale.

Recent advances in detection and tracking using CNNs \cite{RCNN,R-FCN,WarpNet} has the potential to augment healthcare providers by (1) detecting points of malignancy, and (2) finding corresponding lesions across images, allowing them to be tracked temporally.
However, the primary challenge in using traditional detection techniques is working in a low-data regime without the availability of high volumes of annotated and labeled data - the largest existing open-source skin cancer dataset of photographic images is the Edinburgh Dermofit dataset, containing 1,300 biopsied images.
To overcome this we develop a domain-specific data synthesis technique which stitches small single-lesion images onto large body images.
Both the body images and the skin lesions images are heavily augmented with various techniques, and the lesions are blended onto the bodies using Poisson image editing \cite{PoissonImageEditing}. 

For large-scale lesion detection, we use this synthetic data to train a fully-convolutional CNN on the task of pixel-wise classification between three classes: background, benign lesion, malignant lesion, then post-processing to generate region proposals.
For image-to-image tracking, a network is trained (using image pairs containing altered lesion and pose positions) to output pixel-wise positional shift.

Our method demonstrates a working end-to-end CNN system capable of tackling two critical diagnostic tasks with superior performance to baseline techniques. It is trained with very little original data and thus the techniques demonstrated here can be easily transferred to other data-limited domains.


\section{Data Synthesis}
We create a domain-specific data augmentation technique for generating synthetic images from two low-data sources: high-quality lesion images (1,300 biopsy-proven cancers and moles), and body images (400 back, leg, and chest images) whose skin regions have been manually segmented.
We first generate images for detection and then further augment them for tracking.
These images are intended to mimic the real-world clinical case of patients exhibiting many lesions, some possibly malignant, with the need to track them over time. 

Detection data is generated in two steps: (1) A blending position on the body image is chosen using local feature matching between a lesion image and the body image, (2) the lesion image is blended into the body image using Poisson image editing.
For tracking, data is generated by further augmenting detection images - that is, given a detection image, we create a pair of images from it.
The purpose here is to recreate temporal images (which will exhibit changes in lesion shape/size, as well as body changes) and to force the network to learn pixel-wise correspondence by focusing on the distortion in local texture information.

\begin{figure}
  \centering
  \includegraphics[width=0.9\textwidth]{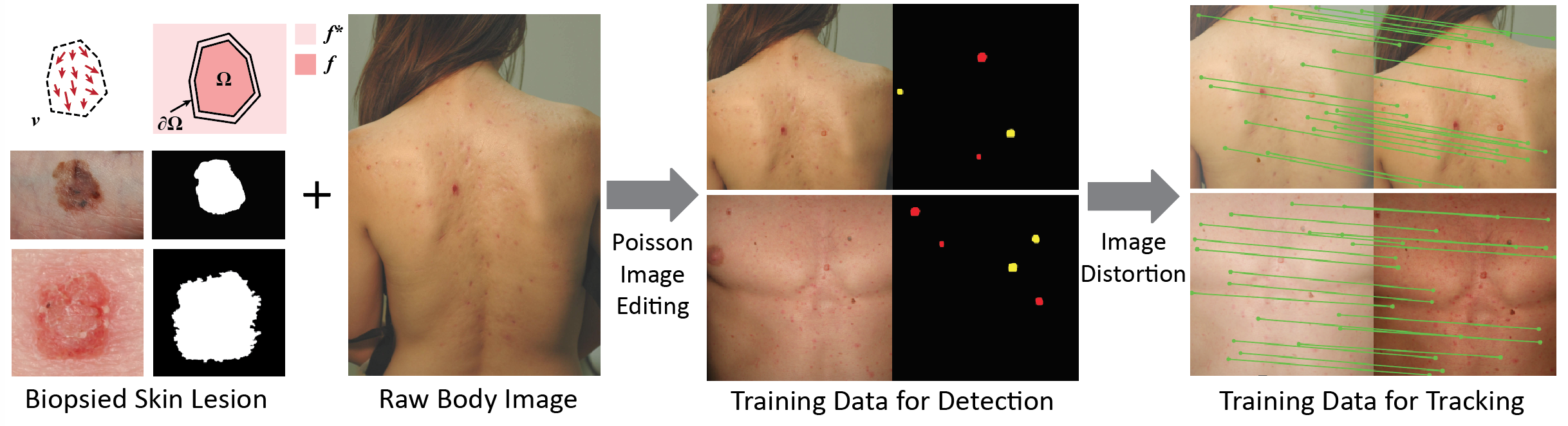}
  \caption{{\bf Data Synthesis.} (Left) Example biopsied skin lesion and raw body images. (Middle) Generated training images for detection and corresponding label masks. Red areas represent blended malignant lesions, yellow areas represent benign lesions. (Right) Generated training images for tracking, along with a few example pixel-wise correspondences.}
  \label{fig:poisson}
\end{figure}

\section{System Pipeline}

\begin{figure*}[h!]
  \centering
  \includegraphics[width=0.9\textwidth]{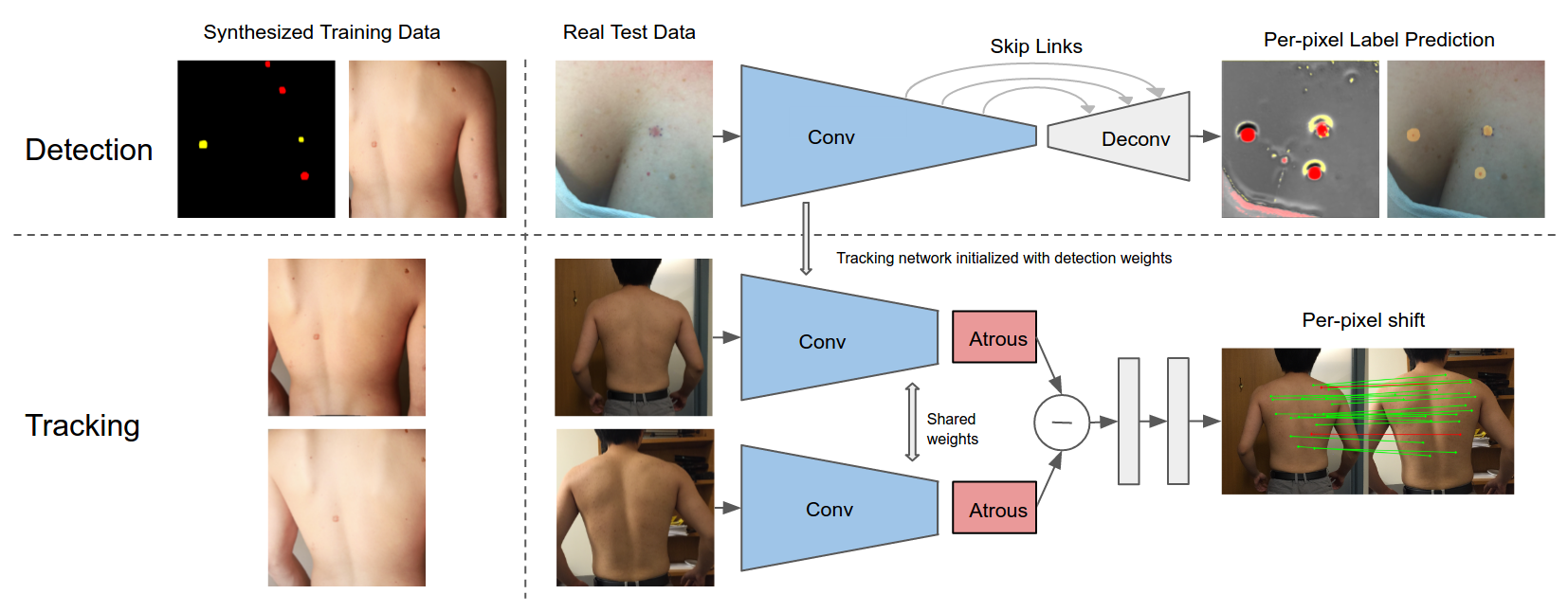}
  \caption{{\bf Detection and Tracking System.} The network is trained on synthetic data and tested on real data. The detection network is composed of a conv section followed by a deconv section, with skip-link connections. In the top right we show the raw prediction heat map and the detection result after post processing. The tracking network takes the conv component of the detection network, and splits it up into a smaller conv part, and an atrous conv part. The two tracking images are each fed through the network and merged by a subtraction before the per-pixel shift prediction.
  }
  \label{fig:system}
\end{figure*}

Our system is composed of two parts: the first detects malignant and benign skin lesions, the second tracks them across images.
The detection network is trained with the synthetic images (skin lesion + body images) described in the previous section, using pixel-wise labels.
Once the detection network is trained to convergence, its weights are used to initialize the tracking network. 
This network is then trained on image-pairs formed from the detection data.

\begin{figure*}[b!]
  \centering
  \includegraphics[width=0.9\textwidth]{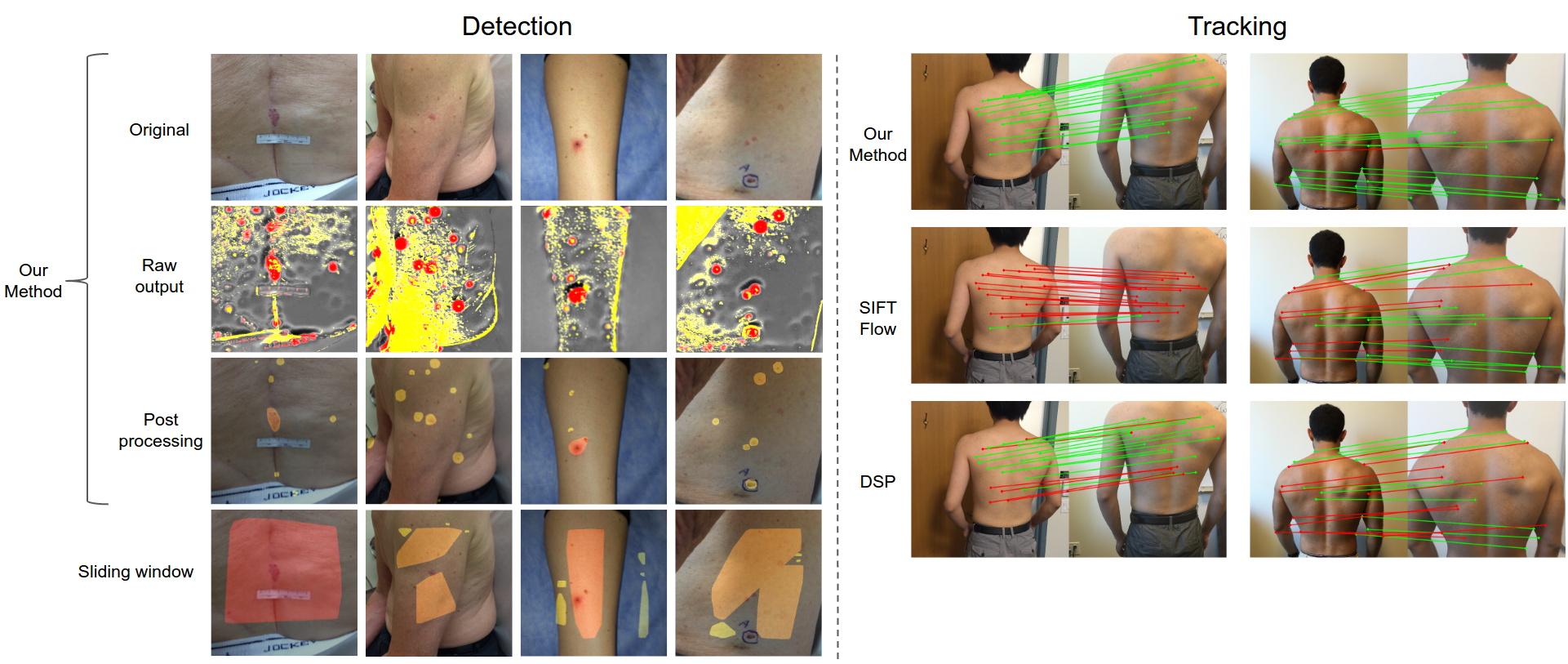}
  \caption{{\bf Qualitative Results.} Left: Detection results, Right: Tracking results. Four examples from our detection pipeline, compared to a baseline sliding-window classifier technique. Two examples from our tracking pipeline, compared to SIFT-Flow and DSP baselines, using $\alpha=0.05$.}
  \label{fig:raw_results}
\end{figure*}

The detection component is intended to highlight to a clinician the potentially malignant lesions on a given input image.
Providers are often faced with body regions containing a multitide of lesions and discerning malignancy is a challenging task. 
Our system feedforwards an input image through the CNN, outputs a pixel-wise heatmap over the five classes of interest, and then uses post-processing techniques to make the heatmap more human-interpretable. 
Examples of raw prediction results and post-processed images are shown in Figure~\ref{fig:raw_results}.

The tracking component of our system is intended to find pixel-wise correspondence between two images of the same body part, in order to track lesions over time. Shown in Figure~\ref{fig:system}, the tracking network is an adaptation of the detection network. During the feedforward pass the two images are fed through the net, which then outputs a 2D vector field of correspondences.


\section{Experiments and Results}

\begin{figure*}
  \centering
  \includegraphics[width=0.3\textwidth]{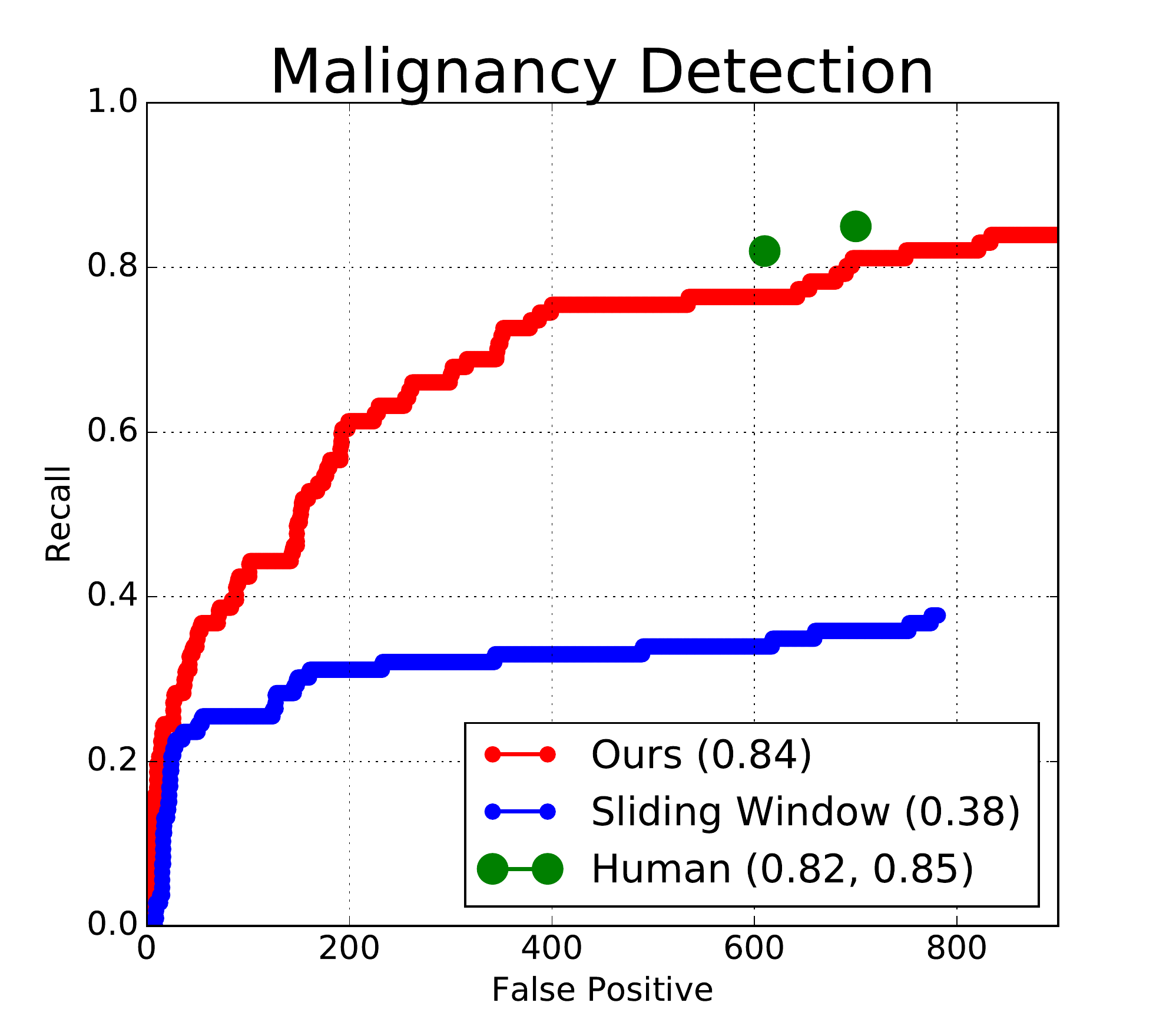}
  \includegraphics[width=0.3\textwidth]{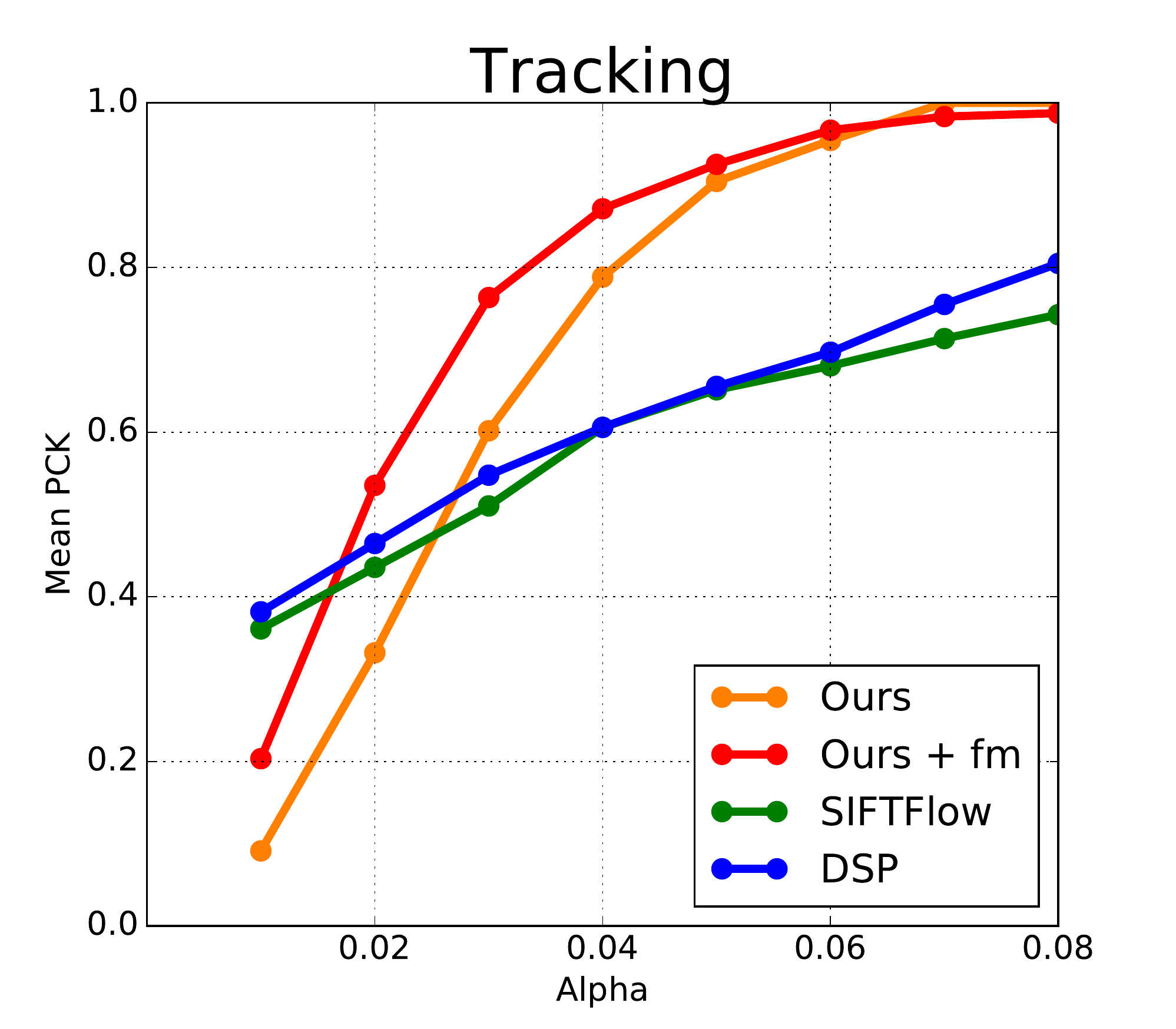}
  \caption{{\bf Quantitative Results.} Detection results (left): ROC curve comparing our technique against a baseline sliding window method and two non-expert humans. Recall rate is shown in the parenthesis of the legend. Tracking Results (right): mean percentage of correct keypoints (PCK) as a function of $\alpha = p/L$, where $p$ is the number of pixels, and $L$ is the diagonal length of the image.}
  \label{fig:quantitative_results}
\end{figure*}


Example results for detection are shown in the first three rows of Figure~\ref{fig:raw_results}, whose columns show examples as they are processed from input images to raw network output to post-processed results. 
Our method is compared to a baseline sliding window method (bottom row of Figure~\ref{fig:raw_results}).
For quantitative comparison, we test detection using 108 clinical images, and include results of two non-expert humans trained on Edinburgh Dermofit dataset to detect malignant lesions.
The generated receiver-operator curve shown in the left part of Figure~\ref{fig:quantitative_results}.

For tracking, we compare our results to two baseline techniques: SIFTFlow, and Deformable Spatial Pyramids (DSP) \cite{SIFTFlow,DeformableSpatialPyramid}.
Example image results are shown in Figure~\ref{fig:raw_results}, where green lines show correctly predicted correspondences, and red lines show incorrect predictions. 
For quantitative comparison, we evaluate tracking accuracy using the percentage-of-correct-keypoints (PCK) metric. Our test set is composed of 260 pairs of correspondence labels manually annotated on temporal image pairs. These image pairs vary in pose, background, distance, viewpoint, and illumination condition.
Results are plotted in Figure~\ref{fig:quantitative_results}, in comparison to both baselines.


\section{Conclusion}

Here we show large-scale detection and tracking of skin lesions across images using FCN in a low-data regime using domain-specific data augmentation.
In the absence of large amounts of labeled and annotated data, we generate high volumes of synthetic data using 1,300 biopsy-proven clinical images of skin lesions and 400 body images. 
Skin lesion images are blended onto body images, heavily augmented with a variety of techniques, and used to train a detection network. 
We demonstrate human-interpretable detection with this method, and demonstrate superior performance over baseline.
We then further augment the data and generate image pairs with pixel-wise correspondence between them, and use this to train a tracking network whose architecture is partially composed of the detection network and initialized with its trained weights, outperforming both SIFTFlow and DSP. 
The networks are trained on synthetic data and tested on real-world data.

AI systems of this sort have the potential to improve the way healthcare is practiced, which may extend outside of the clinic.
Algorithms such as these could aid providers at spotting suspicious lesions amongst benign ones, and at observing temporal changes in lesions that may signify malignancies. 

\small
\bibliography{ref}

\end{document}